\documentclass{article}

\usepackage{PRIMEarxiv}

\usepackage[utf8]{inputenc} 
\usepackage[T1]{fontenc}    
\usepackage{hyperref}       
\usepackage{url}            
\usepackage{booktabs}       
\usepackage{amsfonts}       
\usepackage{nicefrac}       
\usepackage{microtype}      
\usepackage{lipsum}
\usepackage{amsmath}
\usepackage{fancyhdr}       
\usepackage{graphicx}       
\graphicspath{{media/}}     

\title{An Opticalmechanics Framework for Dynamic Estimation of Multibody Systems}

\author{
  Banglei Guan \\
  National University of Defense Technology \\
  \texttt{guanbanglei12@nudt.edu.cn} \\
   \And
  Xuanyu Bai \\
  National University of Defense Technology \\
  \texttt{baixuanyu20@nudt.edu.cn} \\
    \And
  Qingquan Chen* \\
  National University of Defense Technology \\
  \texttt{chenqingquan@nudt.edu.cn} \\
   \And
  Zibin Liu \\
  National University of Defense Technology \\
  \texttt{liuzibin19@nudt.edu.cn} \\
   \And
  Dongcai Tan \\
  National University of Defense Technology \\
  \texttt{tandongcai23@nudt.edu.cn} \\
   \And
  Zhenbao Yu \\
  National University of Defense Technology \\
  \texttt{zhenbaoyu@whu.edu.cn} \\
   \And
  Yang Shang \\
  National University of Defense Technology \\
  \texttt{shangyang1977@nudt.edu.cn} \\
   \And
  Qifeng Yu \\
  National University of Defense Technology \\
  \texttt{yuqifeng@vip.sina.com} \\  
}

\begin{document}
\maketitle

\begin{abstract}
Conventional dynamics analysis of the human body is often constrained by the need for contact force and torque sensors and controlled laboratory environments. To address this issue, this study proposes an opticalmechanics kinematic-dynamic integrated estimation framework for multibody systems. Specifically, a constrained multibody model is established to describe the system dynamics, while image-measured kinematic quantities are used as non-contact inputs for dynamic estimation. The unknown joint torque is then identified through a genetic-algorithm-based optimization by minimizing the discrepancy between model-predicted and image-measured kinematic quantities. Experimental validation on an air-bearing platform showed that the wrist joint torque estimated from image data achieved a mean absolute error of $0.46~\mathrm{N \cdot m}$ compared with sensor measurements. In the forward-prediction test, the model-predicted angular velocity achieved a mean absolute error of $0.006~\mathrm{rad/s}$ relative to the image-measured results. This study demonstrates the potential of combining image measurement and mechanical modeling for non-contact dynamic estimation in scenarios where direct force and torque measurement is difficult.
\end{abstract}


\section{Introduction}
The analysis of human dynamics is fundamental to biomechanics, bionic robotics, and aerospace engineering~\cite{ref1,ref2,ref3}. Accurate quantification of key dynamics parameters, such as external and internal force and torque, is essential for assessing movement intensity and diagnosing injuries~\cite{ref4,ref5}. Traditional high-precision dynamic analysis typically relies on specialized equipment, such as force plates, and on inverse-dynamics methodologies~\cite{ref6,ref7,ref8,ref9}. While these methods provide high accuracy, they are typically implemented in controlled laboratory settings and often rely on specialized equipment, which limits their applicability in extreme or field environments~\cite{ref10}. For example, during astronaut lunar missions and extravehicular activities, it is unfeasible to attach contact sensors to directly measure forces and moments at key body joints. Consequently, the development of non-contact, and easily deployable dynamic analysis methods has emerged as a critical research imperative. In recent years, the convergence of image-based measurements and deep learning techniques has driven significant progress in image-based human dynamics analysis, offering new solutions to address the limitations of traditional approaches~\cite{ref11}.

Early approaches to image-based kinematic estimation depended primarily on marker-based tracking of target objects~\cite{ref12,ref13}. Cao et al.~\cite{ref14} proposed a pose estimation method for effectively extracting 2D joint coordinates directly from images. More recently, advances in deep learning have further improved 3D human pose estimation from images~\cite{ref15,ref16,ref17,ref18}. Pavllo et al.~\cite{ref19} introduced the VideoPose3D model, which accurately reconstructs 3D joint motion trajectories from monocular or multi-view videos. Additionally, Zheng et al.~\cite{ref20} incorporated the Transformer architecture~\cite{ref21} into video-based 3D pose estimation, also demonstrating excellent performance. These studies have established effective approaches for non-contact motion observation and kinematic analysis.

To link image measurement with dynamic analysis, current studies primarily incorporate image information into multibody dynamic model~\cite{ref22}. For example, Shimada et al.~\cite{ref23} proposed PhysCap, which enables real-time 3D human motion estimation from monocular video. Rempe et al.~\cite{ref24} introduced a probabilistic generative model capable of inferring not only 3D pose and shape but also key kinetic parameters. Payton et al.~\cite{ref25} integrated image measurement technology with a multibody dynamic model for motion simulation and inverse dynamics analysis, achieving a combination of image measurement and dynamic analysis. These studies provide a well-established mechanical basis for the analysis of motion in multibody systems.

Building on the above developments in image-based motion measurement and mechanics-based dynamic analysis, this study proposes an opticalmechanics dynamic estimation framework for multibody systems. The central innovation is the establishment of an opticalmechanics framework that combines image-based kinematic measurement with mechanics-based dynamic estimation. Specifically, the contributions of this work are summarized as follows:

\begin{itemize}
\item An opticalmechanics analysis framework is established, in which image-measured kinematic quantities are introduced into a mechanics-based method for non-contact dynamic estimation.
\item A constrained multibody dynamic model is developed, in which equivalent torsional spring constraints are introduced to characterize human joint mechanical behavior and support inverse torque estimation.
\item An optimization-based inverse torque estimation method is proposed, where the unknown joint torque is identified through genetic-algorithm-based optimization and numerical integration.
\end{itemize}

The subsequent sections of this paper are organized as follows: Section~\ref{2} details the fundamental principles and framework. Section~\ref{3} presents the experimental validation and analysis. Section~\ref{4} provides the conclusion.
\section{Proposed methods}
\label{2}
\begin{figure}[t]
\centering
\includegraphics[width=\textwidth]{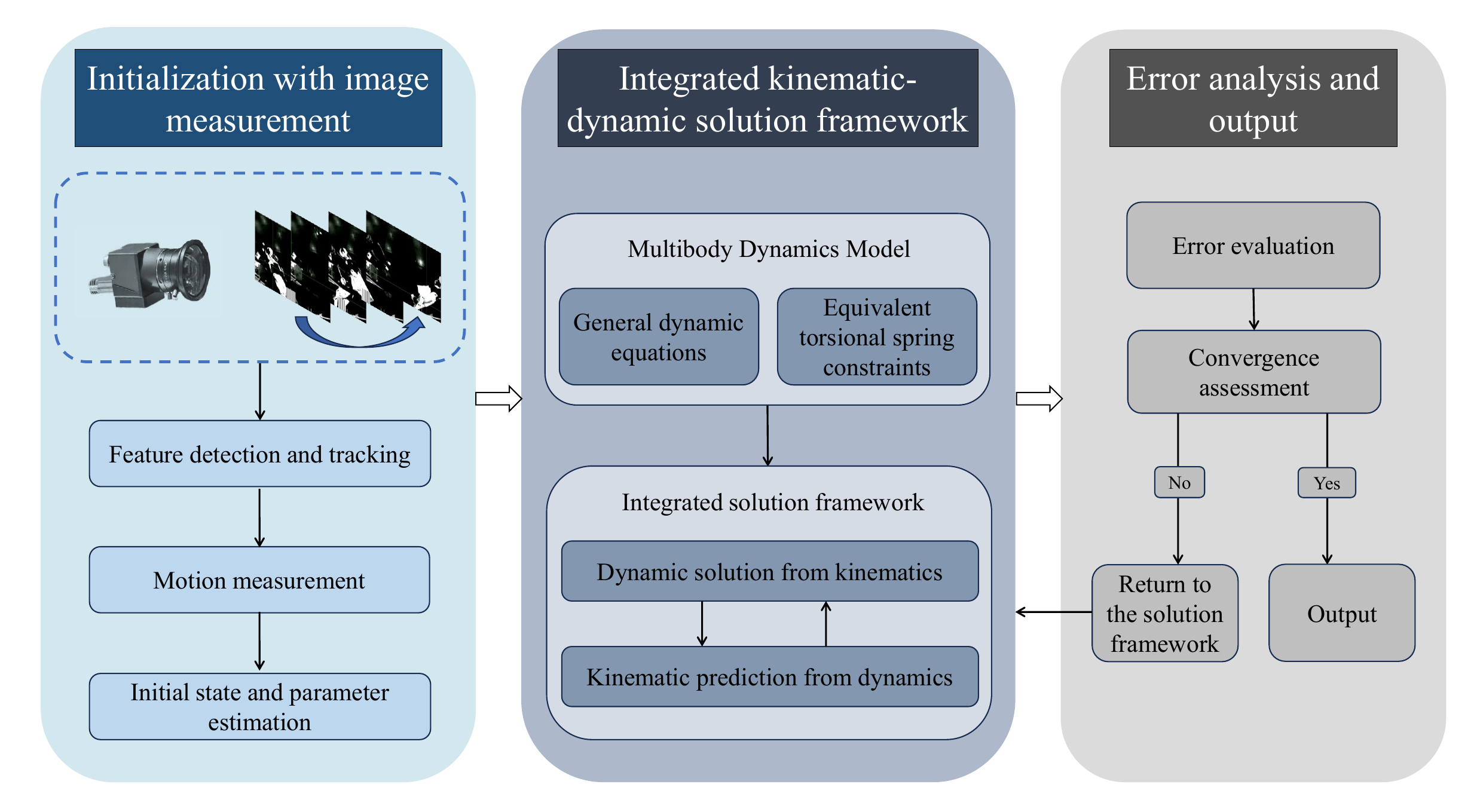}
\caption{Framework of integrated kinematic-dynamic analysis based on image measurement.}
\label{fig1}  
\end{figure}

\noindent Figure~\ref{fig1} depicts the overall opticalmechanics analysis framework. This section presents the proposed method, including optimization-based inverse torque estimation, image-based kinematic measurement, and multibody dynamic modeling. First, an optimization-based inverse torque estimation method is formulated for identifying unknown joint torque. Second, an image-based measurement method is introduced to extract kinematic quantities from image sequences. Third, a constrained multibody model is constructed to provide the mechanical basis for the estimation process. 

\subsection{Optimization-based dynamic estimation method}
\label{2.1}
\noindent To estimate the unknown joint torque in a non-contact manner, an optimization-based inverse dynamic estimation method is first formulated. The core idea is to identify the torque input by minimizing the discrepancy between the model-predicted angular velocity and the image-measured angular velocity at the next time step.

At two consecutive instants $t_i$ and $t_{i+1}$, let $\omega_i^{m}$ and $\omega_{i+1}^{m}$ denote the angular velocities obtained from image measurement. For a candidate torque $M_i$ acting over the interval $[t_i,t_{i+1}]$, the mechanical model predicts the angular velocity at the next instant as
\begin{align}
\omega_{i+1}^{p}=f(M_i,\omega_i^{m}),
\end{align}

\noindent where $f(\cdot)$ represents the dynamic update governed by the multibody model. Accordingly, the inverse estimation of torque is converted into the minimization of the discrepancy between the predicted and measured angular velocities at the next time step. The objective function is defined as
\begin{align}
J(M_i)=\left(\omega_{i+1}^{m}-\omega_{i+1}^{p}\right)^2.
\end{align}

By minimizing $J(M_i)$ in the admissible interval $[M_{\min},M_{\max}]=[-10,10]$, the torque corresponding to the current time interval can be identified as
\begin{align}
M_i^{*}=\arg \min_{M_i \in [M_{\min},M_{\max}]} J(M_i).
\end{align}

The optimization is carried out using a genetic algorithm, which is suitable for the present nonlinear inverse estimation problem. Since the torque at adjacent instants is expected to vary continuously, a prior estimate is introduced to improve convergence efficiency. Specifically, the initial population at the $i-th$ step is generated from the identified torque at the previous step:
\begin{align}
M_i^{(0)}=
\begin{cases}
0, & i=1 \\
M_{i-1}^{*}, & i>1.
\end{cases}
\end{align}

Thus, the initial population matrix is constructed by repeating this prior value over the population members. In this way, the optimization at each step starts from a physically reasonable neighborhood while still allowing the genetic algorithm to search within the prescribed bounded interval. In numerical calculation, the population size is set to $50$ and the maximum number of generations is set to $100$. The function tolerance and the constraint tolerance are both taken as $10^{-6}$. The optimized solution at each time step is recorded as the identified torque sequence.
\begin{align}
\mathbf{M}^{}=\left[M_1^{*},M_2^{*},\dots,M_{N-1}^{*}\right].
\end{align}

In this way, inverse torque identification is converted into a sequential optimization problem constrained by the dynamic model. The measured angular velocity at the current instant serves as the kinematic input, while the unknown torque is determined by minimizing the prediction error at the next instant. The image-measured angular velocity required by this optimization process is obtained through the image-based kinematic measurement method described in Section~\ref{2.2}.
\subsection{Image-based kinematic measurement method}
\label{2.2}
To provide the measurable kinematic quantities required by the torque estimation method, an image-based kinematic measurement procedure based on monocular image sequences is introduced in this section.
\begin{figure}[t]
\centering
\includegraphics[width=\textwidth]{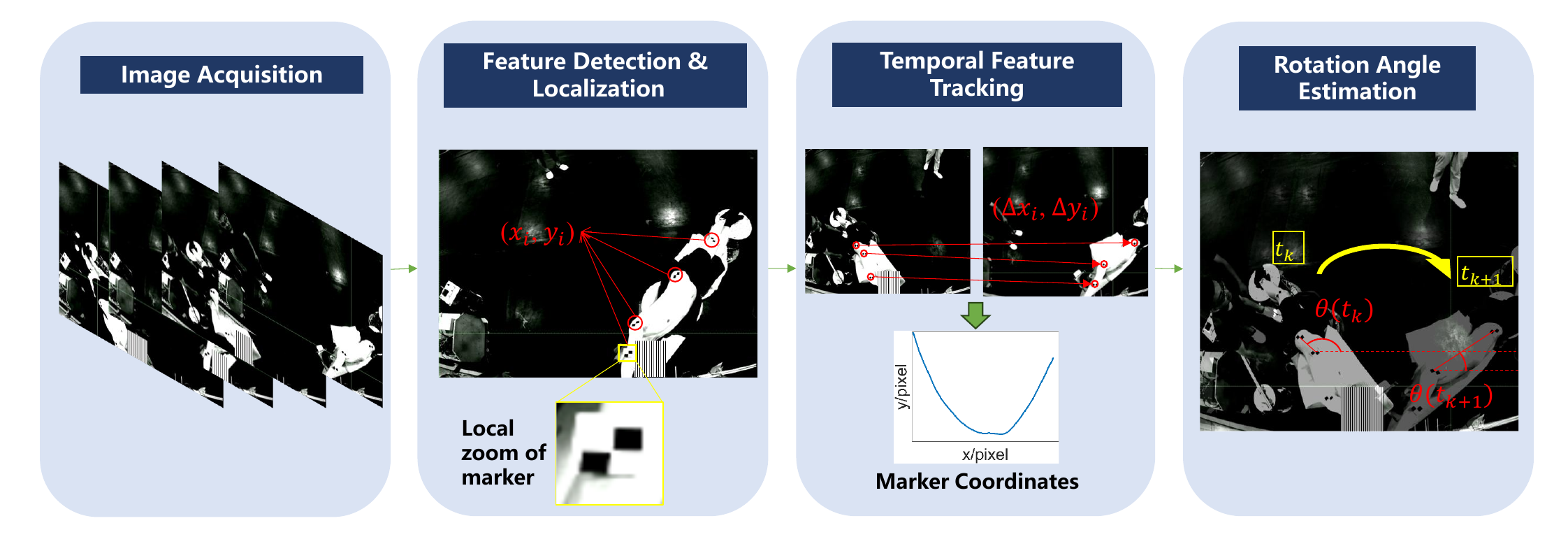}
\caption{Schematic of the image-based kinematic measurement procedure.}
\label{fig5}  
\end{figure}

The overall image-based kinematic measurement procedure is depicted in Fig.~\ref{fig5}. The present method adopts a marker-referenced local scaling strategy. A square paper marker with a cross-shaped pattern and known geometric dimensions is attached to each body segment. The physical side length of each square marker is denoted by $L_0=0.10~\mathrm{m}$. For the $i-th$ marker at time $t_k$, its centroid coordinate in the image plane is written as
\begin{align}
\mathbf{c}_i(t_k)=
\begin{bmatrix}
x_i(t_k)\\
y_i(t_k)
\end{bmatrix},
\end{align}

\noindent where $x_i(t_k)$ and $y_i(t_k)$ are the horizontal and vertical image coordinates, respectively. Meanwhile, let $p_i(t_k)$ denote the side length of the same square marker measured in pixels at time $t_k$. A local scale factor at the marker location is then defined as
\begin{align}
s_i(t_k)=\frac{L_0}{p_i(t_k)}.
\end{align}

This factor represents the physical length corresponding to one pixel in the local region of the $i-th$ marker. 

The wrist is taken as the rotation center throughout the motion. Let the image coordinate of the wrist rotation center at time $t_k$ be denoted by
\begin{align}
\mathbf{c}_0(t_k)=
\begin{bmatrix}
x_0(t_k)\\
y_0(t_k)
\end{bmatrix}.
\end{align}

After local scale normalization, the corresponding position vector in locally normalized physical coordinates for the $i-th$ marker is written as
\begin{align}
\mathbf{r}_i(t_k)=s_i(t_k)[\mathbf{c}_i(t_k)-\mathbf{c}_0(t_k)].
\end{align}

For the $i-th$ marker, the angular displacement between two consecutive frames can be determined from the change in the relative position vector with respect to the wrist rotation center:
\begin{align}
\Delta \theta_i(t_k)=
\arctan
\frac{
\mathbf{r}_i(t_k)\times \mathbf{r}_i(t_{k+1})
}{
\mathbf{r}_i(t_k)\cdot \mathbf{r}_i(t_{k+1})
},
\end{align}

\noindent where $\mathbf{r}_i(t_k)\times \mathbf{r}_i(t_{k+1})$ denotes the scalar out-of-plane component of the two-dimensional cross product, and $\mathbf{r}_i(t_k)\cdot \mathbf{r}_i(t_{k+1})$ denotes the dot product.

Since multiple markers are arranged on the body, the angular velocity of the body is obtained by averaging the angular velocities calculated from the selected markers. $N$ is the number of body markers used in the calculation. For the $i-th$ marker, the angular velocity at time $t_k$ is written as
\begin{align}
\omega_i(t_k)=\frac{\Delta \theta_i(t_k)}{\Delta t},
\end{align}

\noindent where $\Delta t$ is the time interval between two consecutive frames. Then, the averaged angular velocity of the body is expressed as
\begin{align}
\omega(t_k)=\frac{1}{N}\sum_{i=1}^{N}\omega_i(t_k).
\end{align}

Through this procedure, the angular velocity sequence is obtained as the measurable kinematic input for inverse torque estimation. The multibody mechanical model described in Section~\ref{2.3} is then used to relate the measured angular velocity to the corresponding joint torque.
\subsection{Mechanical modeling of the constrained multibody system}
\label{2.3}
\noindent To provide the mechanical basis for the dynamic estimation process described above, a constrained multibody dynamic model is constructed.
\begin{figure}[h]
\centering
  \includegraphics[width=.3\columnwidth]{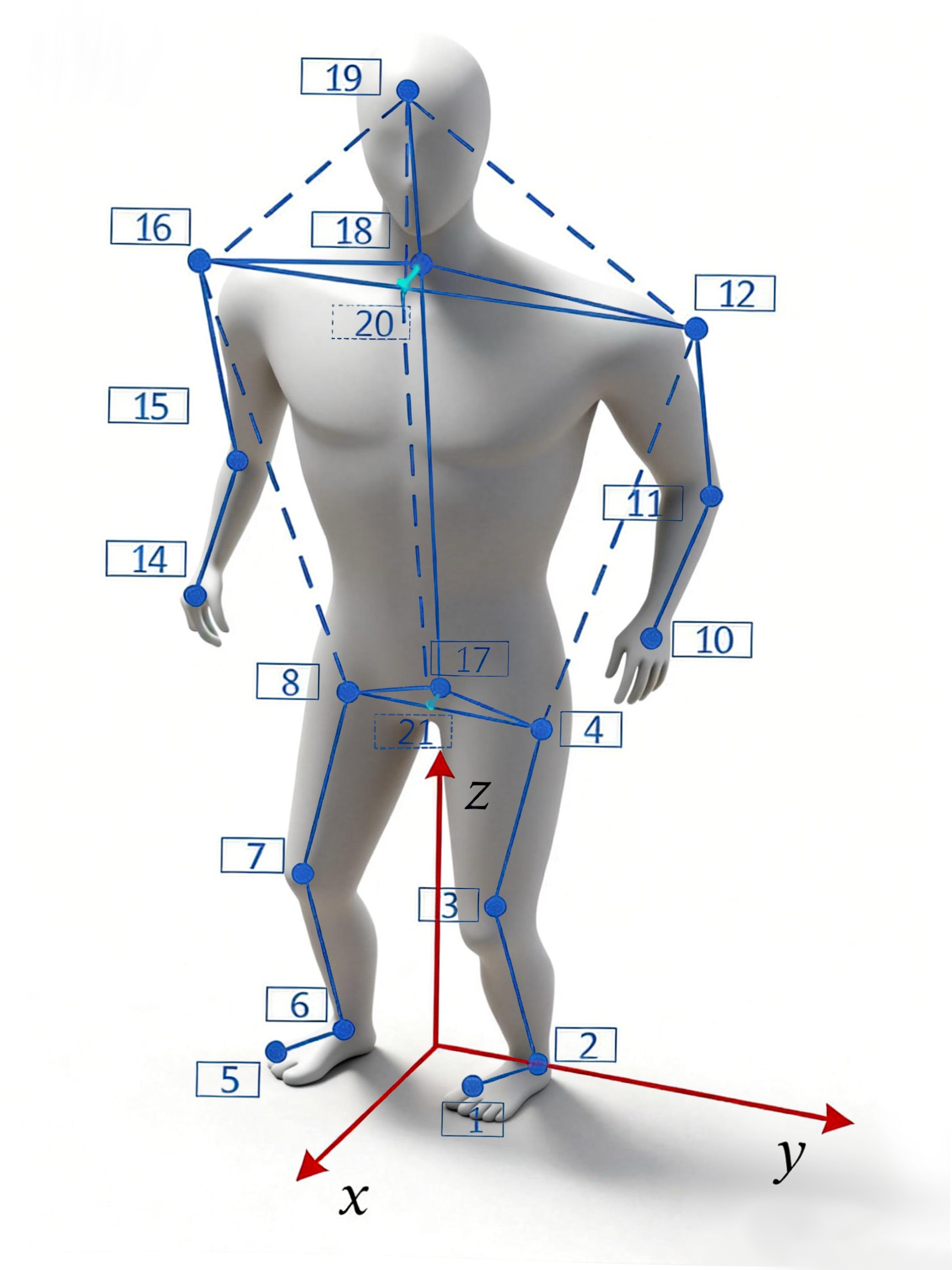}
\caption{Simplified link and node model of the human body.}
\label{fig2}  
\end{figure}

In this model, the body is idealized as a system of rigid segments connected by joints, and each segment is simplified as a rod with uniform mass density for dynamic analysis. As illustrated in Fig.~\ref{fig2}, the human body is idealized as a multibody system comprising 21 nodes and 18 links. In this model, the solid lines represent the physical rigid links with defined mass and moments of inertia, characterizing the main segments of the structure. The dashed lines indicate the topological connections and geometric constraints that define the spatial relationship between adjacent nodes, ensuring the structural consistency of the multibody dynamics framework. Additionally, a faint silhouette is included as a symbolic representation of the human profile, providing a visual context for the mechanical model's orientation and scale.

Within the framework of analytical mechanics, the system's governing equations are established from the principle of virtual displacement and written in the form of constrained Lagrange equations.
\begin{align}
    \mathrm{\frac{d}{dt}}(\frac{\partial T}{\partial\dot{q}_j})-\frac{\partial T}{\partial q_j}=Q_j+\sum_{k=1}^r\lambda_kb_{kj},
\label{eq1}  
\end{align}
\noindent where ${q}_j(j=1,2,\cdots,l)$ denotes the generalized coordinate, $Q_j(j=1,2,\cdots,l)$ is the generalized external force, $T$ is the system's kinetic energy, $\lambda_k$ is the Lagrange multiplier associated with the $k$-th constraint, and $b_{kj}$ is the coefficient relating the $k$-th constraint to the $j$-th generalized coordinate.

During the solution phase, parameters such as link lengths and constraint relationships are embedded into the equations. By formulating dynamic equations with nodal accelerations as unknowns and solving them simultaneously, the accelerations of each node are determined. These are then integrated to compute real-time nodal poses, completing the dynamic solution for the nodal motion states.
\begin{figure*}[t]
\centering
\includegraphics[width=\textwidth]{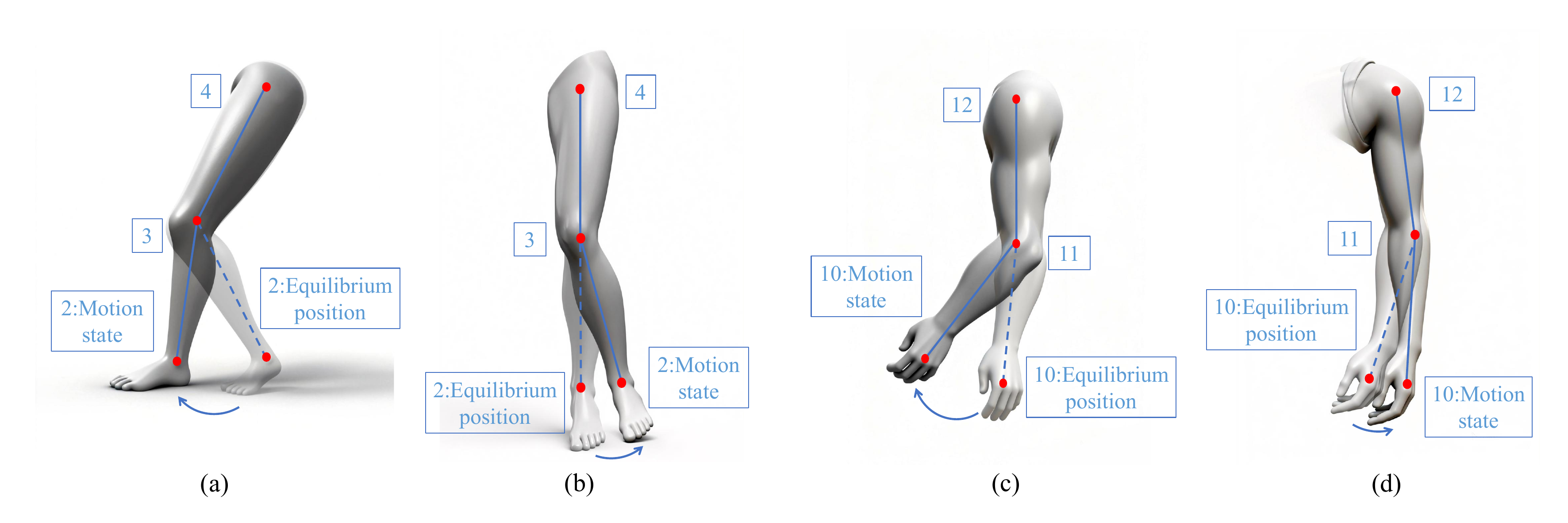}
\caption{Kinematic modeling and motion analysis of distal hinge joints. The dashed contour and solid contour represent the equilibrium position and motion state respectively. The blue arrows indicate the direction of limb movement. The blue dashed boxes represent the simplified model nodes. (a) Knee Flexion and Extension (side view); (b) Knee Abduction and Adduction (front view); (c) Elbow Flexion and Extension (side view); (d) Elbow Abduction and Adduction (front view).}
\label{fig3}  
\end{figure*}

\begin{figure*}[t]
\centering
\includegraphics[width=\textwidth]{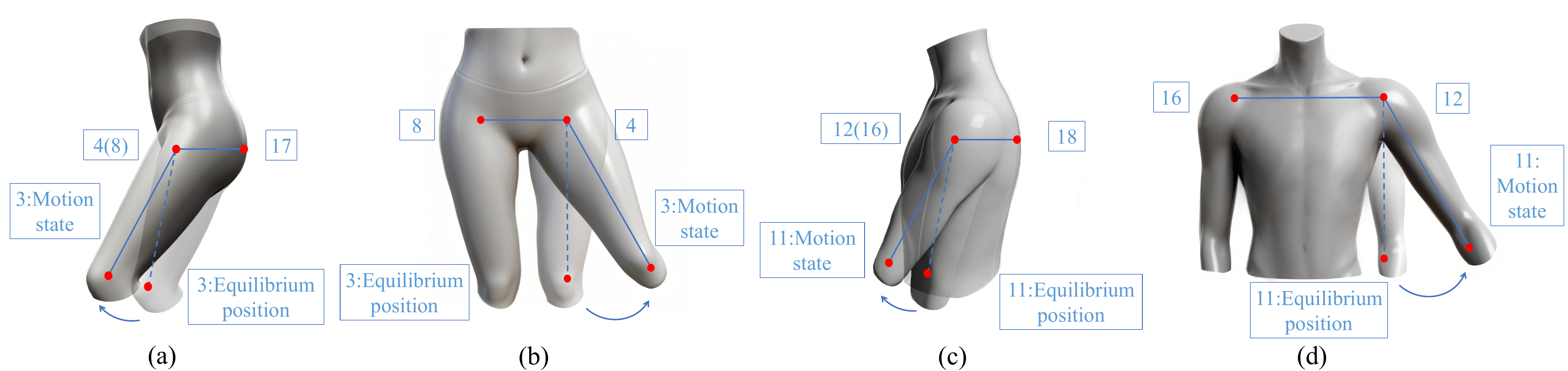}
\caption{Kinematic modeling and motion analysis of proximal complex joints relative to the trunk. (a) Hip Flexion and Extension (side view); (b) Hip Abduction and Adduction (front view); (c) Shoulder Flexion and Extension (side view); (d) Shoulder Abduction and Adduction (front view). Only half-body models are shown in (b) and (d) to improve visual clarity.}
\label{fig4}  
\end{figure*}

Furthermore, to simulate the mechanical behavior of internal joints, an equivalent torsional spring constraint is introduced, facilitating the transmission of joint moments within the dynamic model. This constraint models the internal joint moment as a function of the relative angular displacement between the equilibrium position and the current motion state. The equilibrium position denotes the stable state of a joint under no external force, while the motion state describes the dynamic condition where the joint deviates from equilibrium due to external loads. As the joint deviates from its equilibrium, restorative internal forces and moments are generated to pull the system back toward its initial configuration. This moment is applied as an equivalent force couple on the end nodes of the adjacent links.

As illustrated respectively in Fig.~\ref{fig3} and~\ref{fig4}, we categorize the joints within the link and node model into two distinct mechanical models based on their constraints: distal hinge joints and proximal complex joints. 

\textbf{Distal hinge joints}, represented by the knee and elbow, are modeled as simplified hinge structures connecting adjacent limb segments as depicted in Fig.~\ref{fig3}. Their motion analysis focuses on two primary dimensions: Flexion/Extension (sagittal plane movement) and Abduction/Adduction (frontal plane movement, such as varus and valgus). In these joints, the mechanical chain is relatively independent, where the internal moment is directly governed by the relative angle between two adjacent rigid links.

\textbf{Proximal complex joints}, including the hip and shoulder, serve as the critical interface between the limbs and the central trunk as shown in Fig.~\ref{fig4}. Similar to distal joints, their motion is also resolved into Flexion/Extension and wide-range Abduction/Adduction. The internal moments are resolved by considering the limb's vector relative to the trunk's reference frame, reflecting the stabilization role these joints play in maintaining core posture during dynamic limb movement.

To clarify the mathematical implementation of these constraints, we provide a detailed analysis of the knee joint as a representative example of how the equivalent torsional spring is applied to resolve joint moments. Figure~\ref{fig3}(a) demonstrates this process for the flexion and extension of the left knee, formed by node 2 (ankle), node 3 (knee), and node 4 (hip). 

First, calculate the internal moment $\boldsymbol{M}_{234}$ generated at the left knee joint due to the flexion movement:
\begin{align}
    \boldsymbol{M}_{234}=\mathrm{K_{234}}\cdot(\theta_{234}-\theta_{234}^0),
\end{align}

\begin{align}
    \theta_{234}=\pi-\arccos\left(\frac{\boldsymbol{n}_{34}\cdot \boldsymbol{n}_{24}}{|\boldsymbol{n}_{34}||\boldsymbol{n}_{24}|}\right)
    -\arccos\left(\frac{\boldsymbol{n}_{23}\cdot
    \boldsymbol{n}_{24}}
    {|\boldsymbol{n}_{23}||\boldsymbol{n}_{24}|}\right),
\end{align}  
\noindent where $K_{234}$ represents the equivalent torsional stiffness of the joint, $\theta_{234}^0$ is the initial angle of the triangle formed by nodes 2-3-4 in the equilibrium position, while $\theta_{234}$ denotes the real-time angle during motion. $\boldsymbol{n}_{23}$ represents the directional vector from node 2 to node 3, which in the model corresponds to the vector pointing from the ankle to the knee. The meanings of the remaining vectors follow the same logic.

Subsequently, this internal moment is converted into a force couple acting on the adjacent links. The force $\boldsymbol{F}_2$ applied at node 2 (ankle), force $\boldsymbol{F}_4$ at node 4 (hip), and force $\boldsymbol{F}_3$ at node 3 (knee) are derived as follows:
\begin{align}
    \boldsymbol{F}_2=\frac{\boldsymbol{M}_{234}}{\left|\boldsymbol{n}_{32}\right|}\cdot \frac{\boldsymbol{n}_{32}\times\left(\frac{\boldsymbol{n}_{32}\times\boldsymbol{n}_{34}}{\left|\boldsymbol{n}_{32}\times\boldsymbol{n}_{34}\right|}\right)}{\left|\boldsymbol{n}_{32}\times\left(\frac{\boldsymbol{n}_{32}\times\boldsymbol{n}_{34}}{\left|\boldsymbol{n}_{32}\times\boldsymbol{n}_{34}\right|}\right)\right|},
\end{align}
\begin{align}
    \boldsymbol{F}_4=\frac{\boldsymbol{M}_{234}}{\left|\boldsymbol{n}_{34}\right|}\cdot
    \frac{\left(\frac{\boldsymbol{n}_{32}\times\boldsymbol{n}_{34}}{\left|\boldsymbol{n}_{32}\times\boldsymbol{n}_{34}\right|}\right)\times\boldsymbol{n}_{34}}{\left|\left(\frac{\boldsymbol{n}_{32}\times\boldsymbol{n}_{34}}{\left|\boldsymbol{n}_{32}\times\boldsymbol{n}_{34}\right|}\right)\times\boldsymbol{n}_{34}\right|},
\end{align}
\begin{align}
    \boldsymbol{F}_3=-\boldsymbol{F}_2-\boldsymbol{F}_4.
\end{align}

In this way, the constrained multibody model establishes the mechanical relationship between motion response and internal joint moments. Together with the image-based kinematic measurement and the optimization strategy, it completes the proposed opticalmechanics framework for non-contact torque estimation.
\section{Experiments}
\label{3}
\begin{figure}[b]
\centering
\includegraphics[width=.75\columnwidth]{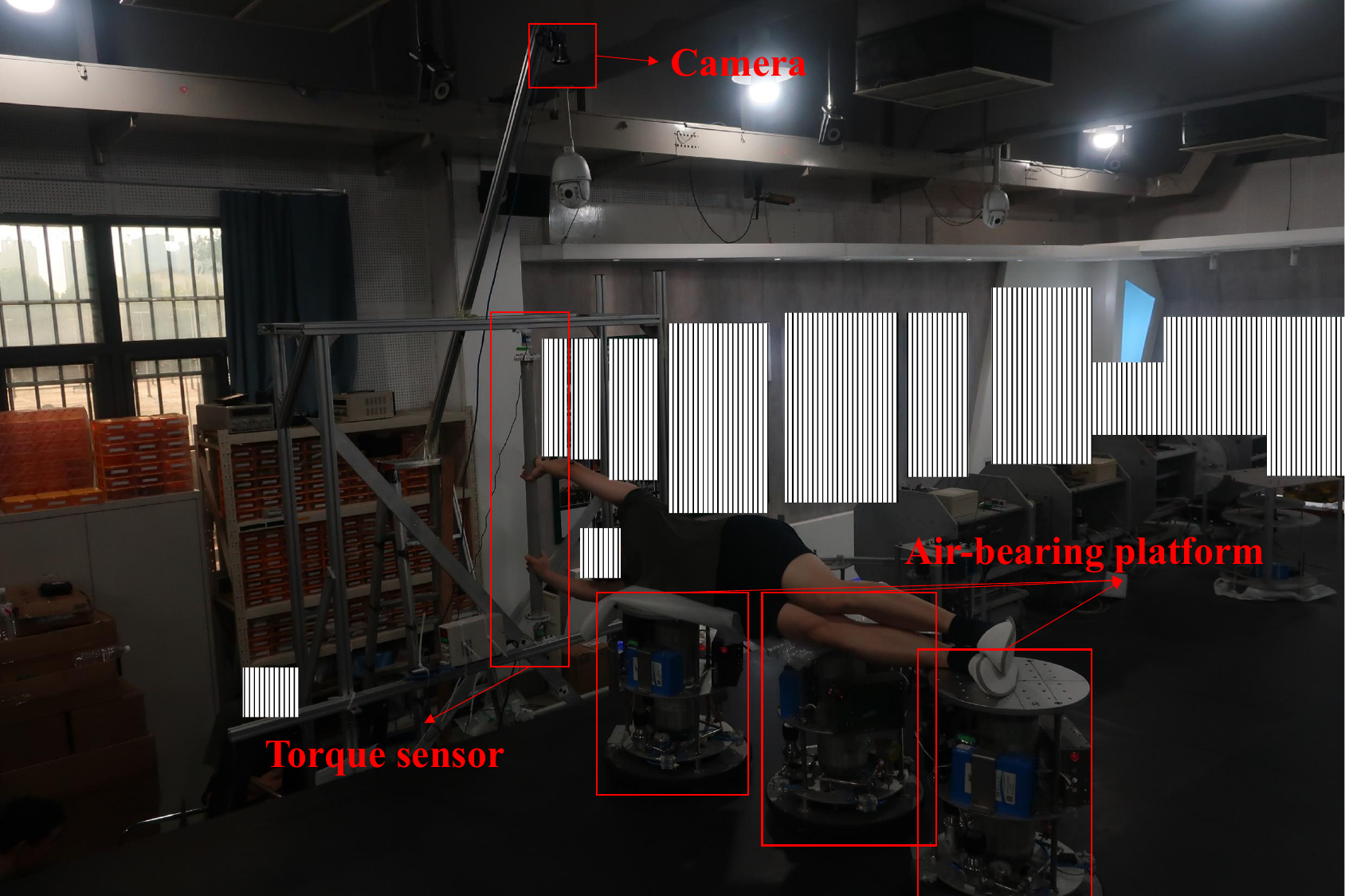}
\caption{Overall experimental configuration with an air-bearing platform.}
\label{fig6}
\end{figure}

To assess the accuracy of the proposed framework, this section outlines a bidirectional validation methodology employing an air-bearing platform. Section~\ref{3.1} details the experimental setup and body parameters of our volunteer. Section~\ref{3.2} validates the dynamic model parameters through forward prediction, while Section~\ref{3.3} evaluates the robustness of the optimization framework in inverse estimation.
\subsection{Experimental environment and configuration}
\label{3.1}
The experiments were conducted on an air-bearing platform, which generates a stable air film to minimize friction between the platform and its surface, effectively simulating a microgravity environment. Moreover, this setup reduces frictional disturbance and facilitates planar rotational motion for controlled validation. A torque sensor (accuracy: $0.2~\mathrm{N \cdot m}$) was mounted on one side, and a high-resolution camera (resolution: 2448 × 2048 \text{pixels}, frame rate: 36 \text{fps}) was positioned overhead to monitor the motion in real-time. The experimental setup is illustrated in Fig.~\ref{fig6}.

\begin{figure}[ht]
\centering
\includegraphics[width=.65\columnwidth]{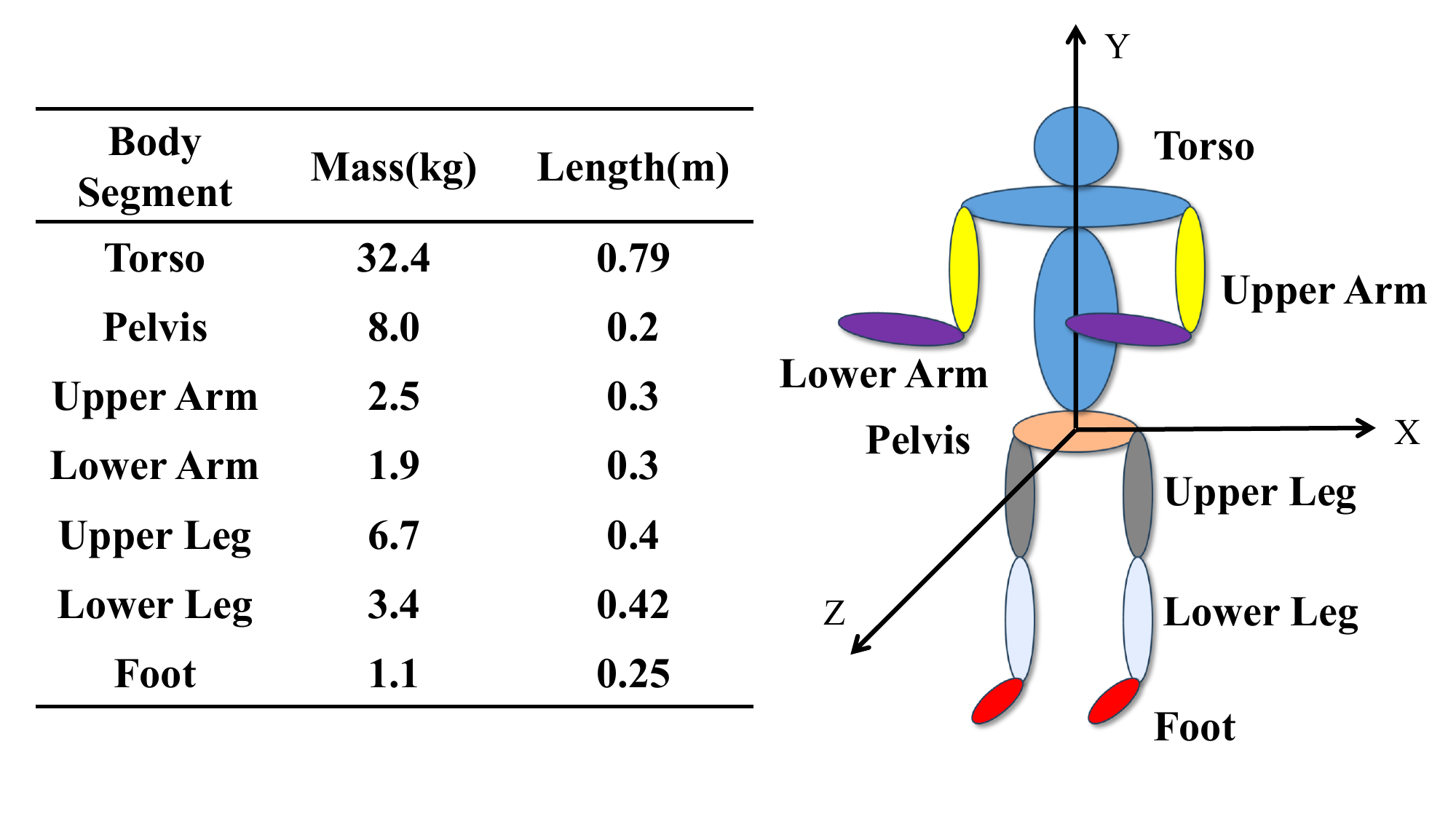}
\caption{Model masses and lengths for body segments of the volunteer in the experiment.}
\label{fig7}
\end{figure}

A volunteer performed rotational movements while holding a rod with both hands. The body parameters of the volunteer are shown in Fig.~\ref{fig7}. Due to the simulated weightless environment, the volunteer's body could be approximately considered a unified rigid body rotating around the wrist, with all body segments sharing an identical angular velocity at each instant. The torque sensor acquired real-time wrist torque data, while the camera captured image sequences of the motion process. 

Cross-shaped markers affixed to key joint locations served as features for the tracking algorithm. This configuration ensured the acquisition of marker's 2D motion trajectories, providing the kinematic inputs required for subsequent dynamic estimation.
\begin{figure}[ht]
\centering
\includegraphics[width=.7\columnwidth]{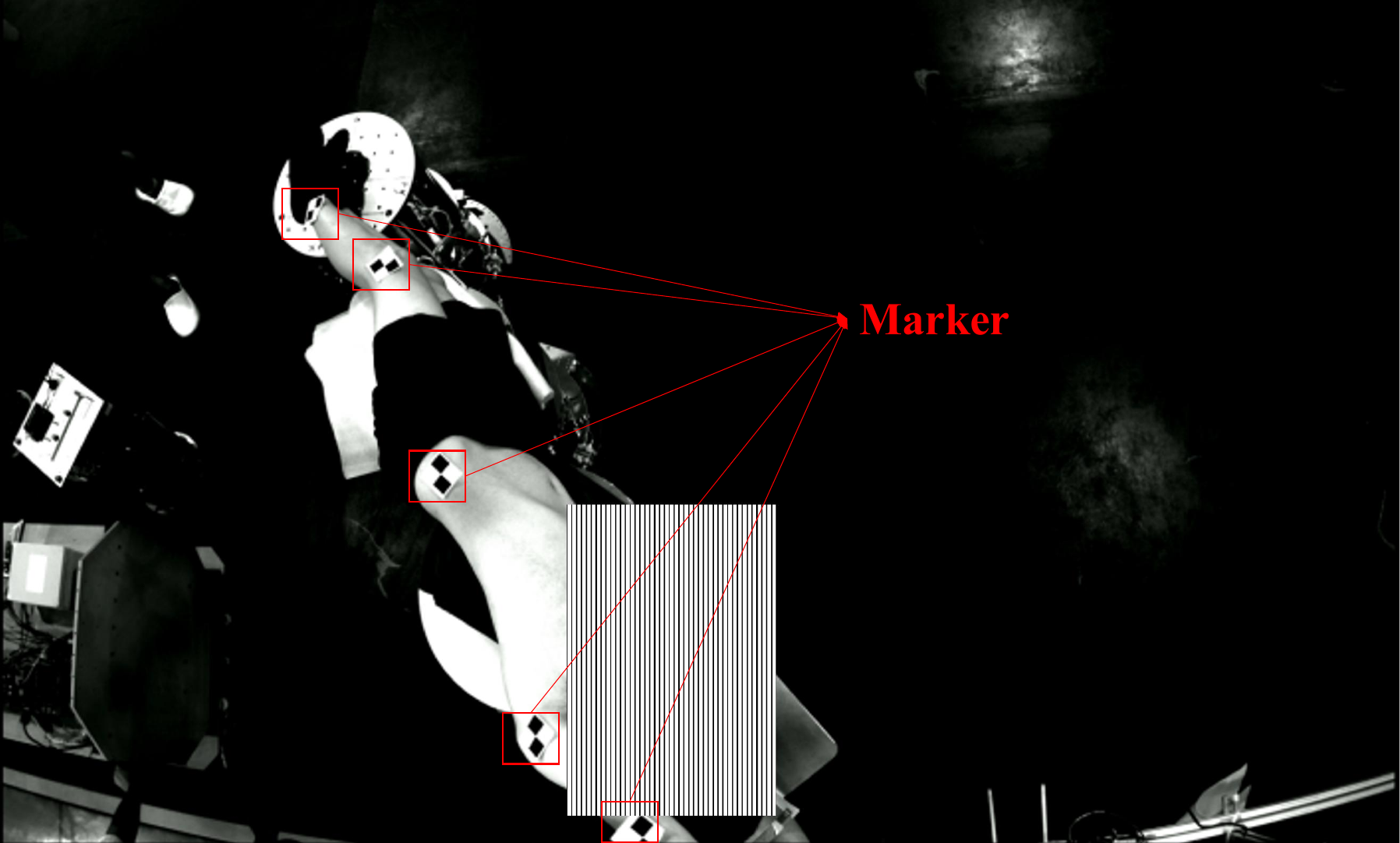}
\caption{Example frames of the camera-captured human motion with attached markers.}
\label{fig8}
\end{figure}

\subsection{Motion angular velocity prediction} 
\label{3.2}
\noindent This section conducts a forward dynamics analysis to validate the predictive capability of the proposed framework. The analysis utilizes reference values from the torque sensor and initial angular velocities derived from image measurements as input for the multibody dynamics model. Through numerical iteration, the model computes the complete angular velocity profile throughout the entire motion. The predicted velocities are then compared against reference data obtained through image measurements.

\begin{figure}[ht]
\centering
  \includegraphics[width=.5\columnwidth]{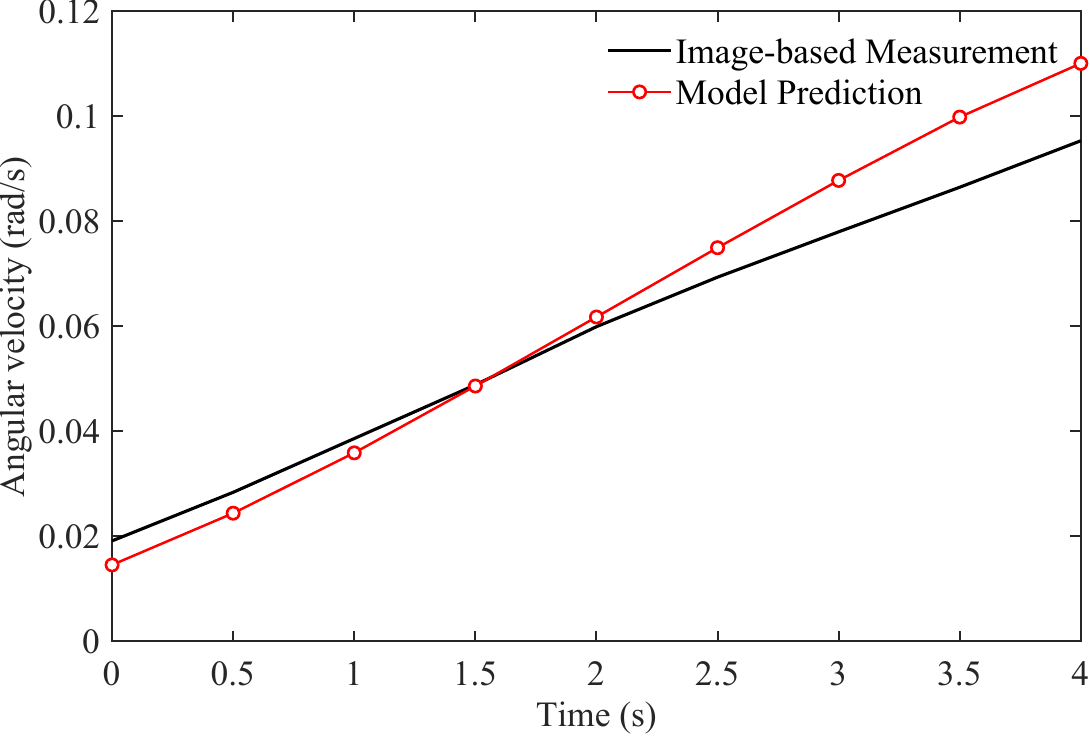}
\caption{Comparison of the model predicted angular velocity and image-based measurements.}
\label{fig9}  
\end{figure}

Figure~\ref{fig9} illustrates the comparison between the model-predicted angular velocities and the image-measured values. The curves demonstrate a high degree of consistency in trend, confirming the model's ability to predict motion based on external torques over the duration of the movement.

\begin{figure}[h]
\centering
\includegraphics[width=0.7\textwidth]{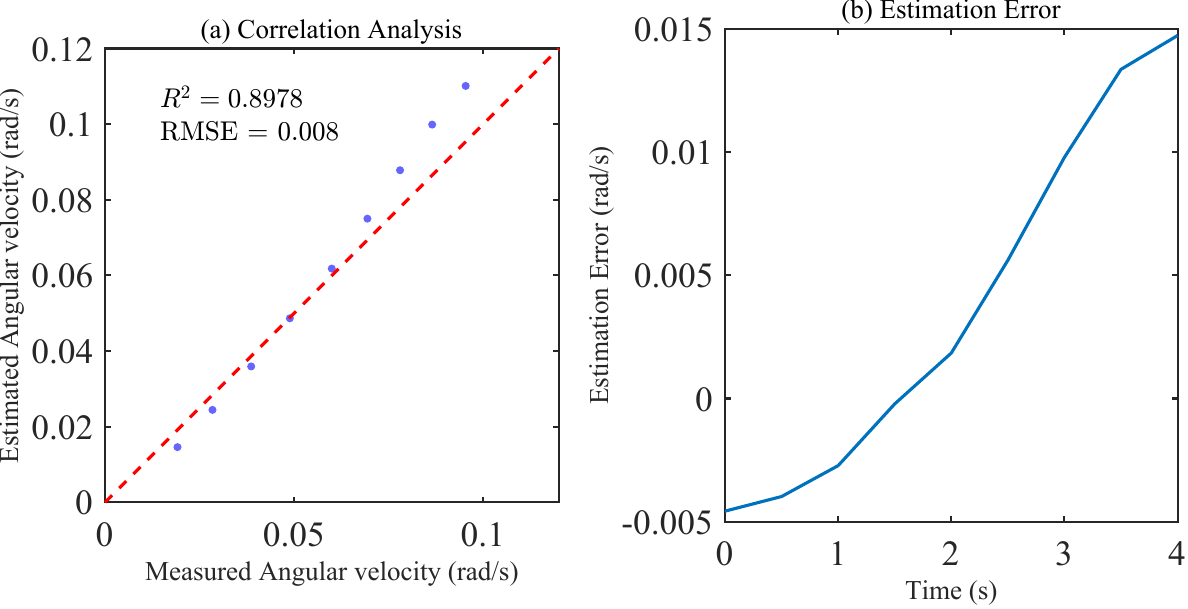}
\caption{Error analysis of the angular velocity prediction.}
\label{fig10}
\end{figure}

Statistical analysis based on Fig.~\ref{fig10}(a) reveals a strong linear correlation between the model-predicted angular velocities and the image-based measurements, characterized by a coefficient of determination ($R^2$) of 0.8978. Fig.~\ref{fig10}(b) indicates that the error remains within a narrow range of $-0.005$ to $0.015~\mathrm{rad/s}$. The resulting mean absolute error (MAE) and root mean square error (RMSE) are $0.006~\mathrm{rad/s}$ and $0.008~\mathrm{rad/s}$, respectively, indicating that the predicted response agrees well with the measured kinematic output. Therefore, the agreement observed in Fig.~\ref{fig9} indicates that the adopted dynamic formulation, parameter setting, and numerical integration procedure can reproduce the principal kinematic behavior of the tested system. 

Meanwhile, the residual discrepancy also provides useful information about the limitations of the current framework. Although the forward prediction is accurate, small deviations still exist, especially near time instants where the angular velocity changes more rapidly. These deviations are likely associated with the uncertainty of marker localization and numerical differentiation involved in the extraction of angular velocity. In addition, the simplified multibody representation and equivalent joint modeling may introduce minor model mismatch. Nevertheless, as the errors remain small throughout the test, the forward-validation results support the view that the proposed opticalmechanics framework is able to provide a reliable mechanical description of the motion and can serve as a basis for subsequent inverse torque estimation.
\subsection{Wrist torque estimation}
\label{3.3}
\noindent This section performs inverse estimation to infer dynamic parameters from the measured kinematic data. Specifically, body angular velocities are extracted through image processing. These measured velocities are fed into the dynamic model, and the optimization framework described in Section~\ref{2.1} is utilized to resolve the wrist torque. Finally, the optimized torque is compared with the reference torque measured by the sensor.
\begin{figure}[ht]
\centering
  \includegraphics[width=0.5\columnwidth]{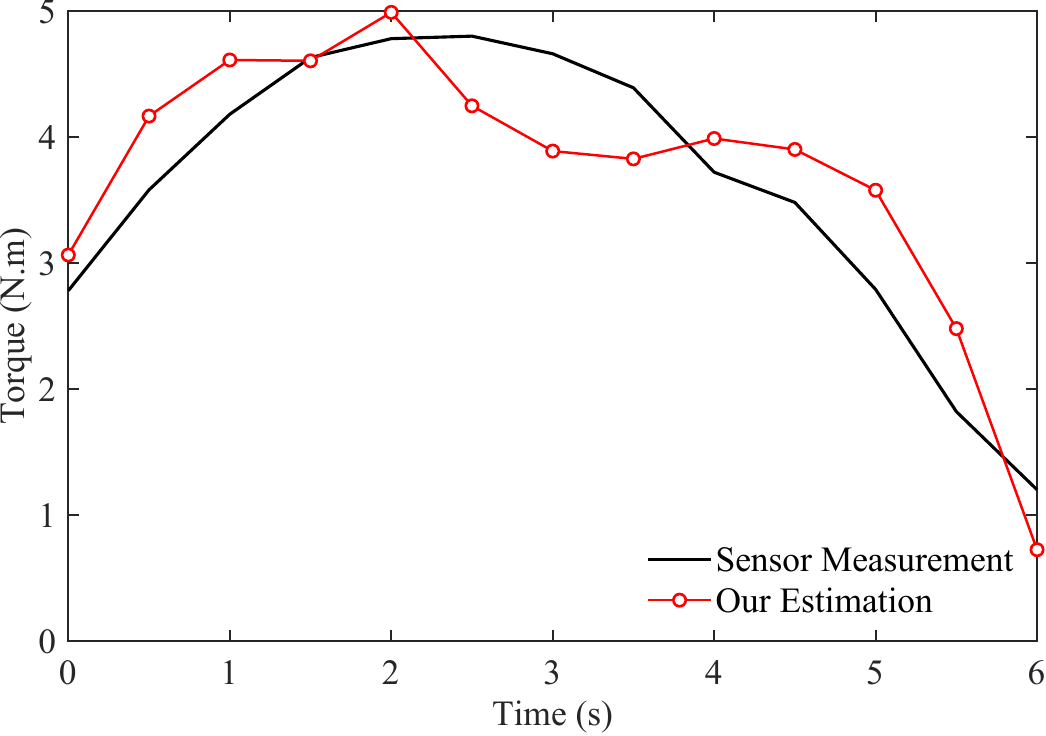}
\caption{Comparison between the estimated wrist torque and sensor-measured torque.}
\label{fig11}  
\end{figure}

\begin{figure}[ht]
\centering
\includegraphics[width=0.7\textwidth]{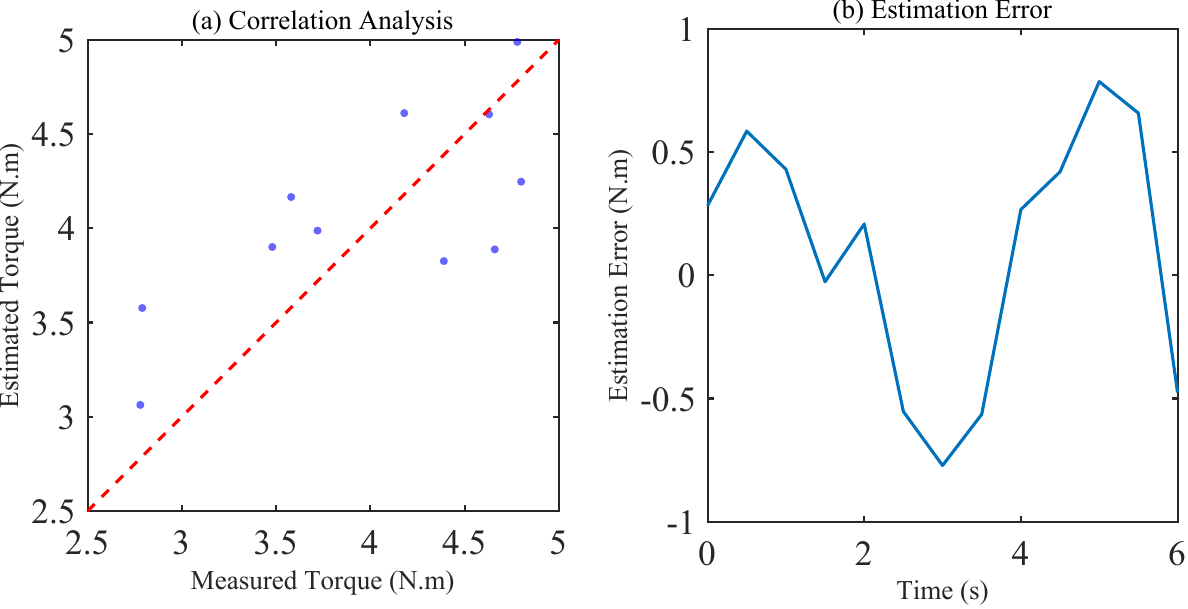}
\caption{Error analysis of the estimated wrist torque.}
\label{fig12}
\end{figure}

The inverse-estimation results demonstrate that the proposed method is capable of reconstructing joint torque from image-measured angular velocity with reasonable overall accuracy, as depicted in Fig.~\ref{fig11}, although the discrepancy is more pronounced than that in the forward validation. As shown in Fig.~\ref{fig12}(a), the estimated torque curve can reproduce the overall loading level and the major temporal fluctuation of the measured signal. Fig.~\ref{fig12}(b) further shows that the error is mainly concentrated within the range of $-0.77~\mathrm{N \cdot m}$ to $0.78~\mathrm{N \cdot m}$, and the corresponding MAE and RMSE are $0.46~\mathrm{N \cdot m}$ and $0.51~\mathrm{N \cdot m}$, respectively. 

These results indicate that the proposed opticalmechanics framework is able to provide a reasonable non-contact approximation of joint torque, although its pointwise accuracy is still lower than that of the forward angular-velocity prediction in Section~\ref{3.2}. These discrepancies are mainly related to simplifications in the multibody mechanical model and the accumulation of local deviations in the sequential optimization process. Nevertheless, despite these error sources, the consistency in the overall torque trend still supports the feasibility of estimating representative dynamic parameters from image-measured kinematic quantities.
\section{Conclusions}
\label{4}
\noindent This paper proposes an opticalmechanics framework for non-contact dynamic estimation of multibody systems. By combining image-based kinematic measurement with mechanics-based dynamic modeling, the framework enables the estimation of joint torque from image sequences throughout the motion.

Experimental validation demonstrated good agreement in forward kinematic prediction and reasonable consistency in inverse torque estimation between the proposed framework and sensor-based measurements. These results demonstrate the feasibility of using image-measured kinematic quantities for non-contact dynamic estimation. The present study provides a practical route for estimating dynamic quantities when conventional contact force and torque measurement equipment is difficult to deploy. Potential applications include astronaut extravehicular activities, sports motion assessment, and robot-related dynamic analysis.

\section*{Acknowledgments}
This work was supported by the National Natural Science Foundation of China (Grant No. 12372189) and the Science and Technology Innovation Program of Hunan Province (Grant No. 2025RC1045).

\section*{Disclosures}
On behalf of all authors, the corresponding author states that there is no conflict of interest.



\begin{thebibliography}{99}
\bibitem{ref1} Kang W, Li L, Song H, et al. Advances and challenges in the mechanics of biological soft tissues: test methods, modeling, and applications. Acta Mech. Sin. 41, 624409, 2025.
\bibitem{ref2} Robertson D G E, Caldwell G E, Hamill J, et al. Research methods in biomechanics. Human Kinetics, 2013.
\bibitem{ref3} Wang Q, Rao Y. Visual Analysis of Human Motion: A Survey on Recent Advances and Applications. IEEE Visual Communications and Image Processing, pp. 1-4, 2018.
\bibitem{ref4} Zatsiorsky V M. Kinetics of human motion. Human Kinetics, 2002.
\bibitem{ref5} Engebretsen L, Wijdicks C A, Anderson C J, Westerhaus B, LaPrade R F. Evaluation of a simulated pivot shift test: a biomechanical study. Knee Surg. Sports Traumatol. Arthrosc. 20(4), 698-702, 2012.
\bibitem{ref6} Winter D A. Biomechanics and Motor Control of Human Movement, 2009.
\bibitem{ref7} Beckham G, Suchomel T, Mizuguchi S. Force Plate Use in Performance Monitoring and Sport Science Testing. New Stud. Athl. 29, 25-37, 2014.
\bibitem{ref8} Oh S E, Choi A, Mun J H. Prediction of ground reaction forces during gait based on kinematics and a neural network model. J. Biomech. 46(14), 2372-2380, 2013.
\bibitem{ref9} Wang X, Guo J, Tian Q. A forward-inverse dynamics modeling framework for human musculoskeletal multibody system. Acta Mech. Sin. 38, 522140, 2022.
\bibitem{ref10} Kanko R M, Laende E K, Davis E M, Selbie W S, Deluzio K J. Concurrent assessment of gait kinematics using marker-based and markerless motion capture. J. Biomech. 127, 110665, 2021.
\bibitem{ref11} Niu Z, Lu K, Xue J, Qin X, Wang J, Shao L. From Methods to Applications: A Review of Deep 3D Human Motion Capture. IEEE Trans. Circuits Syst. Video Technol. 34(11), 11340-11359, 2024.
\bibitem{ref12} Agarwal A, Triggs B. 3D human pose from silhouettes by relevance vector regression. IEEE Computer Society Conference on Computer Vision and Pattern Recognition, 2, II-II, 2004.
\bibitem{ref13} Colyer S L, Evans M, Cosker D P, Salo A I T. A Review of the Evolution of Vision-Based Motion Analysis and the Integration of Advanced Computer Vision Methods Towards Developing a Markerless System. Sports Med. Open 4(1), 24, 2018.
\bibitem{ref14} Cao Z, Hidalgo G, Simon T, et al. OpenPose: realtime multi-person 2D pose estimation using part affinity fields. IEEE Trans. Pattern Anal. Mach. Intell. 43(1), 172-186, 2021.
\bibitem{ref15} Zheng C, et al. Deep Learning-based Human Pose Estimation: A Survey. ACM Comput. Surv. 56, 1-37, 2020.
\bibitem{ref16} Zhao L, Peng X, Tian Y, Kapadia M, Metaxas D N. Semantic Graph Convolutional Networks for 3D Human Pose Regression. IEEE/CVF Conference on Computer Vision and Pattern Recognition, pp. 3420-3430, 2019.
\bibitem{ref17} Choi H, Moon G, Lee K M. Pose2Mesh: Graph Convolutional Network for 3D Human Pose and Mesh Recovery from a 2D Human Pose. European Conference on Computer Vision, VII, 769-787, 2020.
\bibitem{ref18} Lin K, Wang L, Liu Z. End-to-End Human Pose and Mesh Reconstruction with Transformers. IEEE/CVF Conference on Computer Vision and Pattern Recognition, pp. 1954-1963, 2021.
\bibitem{ref19} Pavllo D, Feichtenhofer C, Grangier D, Auli M. 3D Human Pose Estimation in Video With Temporal Convolutions and Semi-Supervised Training. IEEE/CVF Conference on Computer Vision and Pattern Recognition, pp. 7745-7754, 2019.
\bibitem{ref20} Zheng C, et al. 3D Human Pose Estimation with Spatial and Temporal Transformers. IEEE/CVF International Conference on Computer Vision, pp. 11636-11645, 2021.
\bibitem{ref21} Vaswani A, Shazeer N, Parmar N, Uszkoreit J, Jones L, Gomez A N, Kaiser {\L}, Polosukhin I. Attention is all you need. International Conference on Neural Information Processing Systems. Curran Associates Inc., 6000-6010, 2017.
\bibitem{ref22} Kok M, Hol J D, Schön T B. Using inertial sensors for position and orientation estimation. Found. Trends Signal Process. 11(1-2), 1-153, 2017.
\bibitem{ref23} Shimada S, Golyanik V, Xu W, Theobalt C. PhysCap: physically plausible monocular 3D motion capture in real time. ACM Trans. Graph. 39(6), Article 235, 16 pages, 2020.
\bibitem{ref24} Rempe D, Birdal T, Hertzmann A, Yang J, Sridhar S, Guibas L J. HuMoR: 3D Human Motion Model for Robust Pose Estimation. IEEE/CVF International Conference on Computer Vision, pp. 11468-11479, 2021.
\bibitem{ref25} Payton C J, Burden A. Biomechanical Evaluation of Movement in Sport and Exercise. The British Association of Sport and Exercise Sciences Guide, 2017.
\bibitem{ref26} Yu Z, Liu S, Tian Q. Kalman filter based state estimation for the flexible multibody system described by ANCF. Acta Mech. Sin. 41, 524373, 2025.
\bibitem{ref27} Huang H, Shang Y, Guan B, et al. 3D trajectory reconstruction of moving points based on asynchronous cameras. Acta Mech. Sin. 41, 425322, 2025.
\bibitem{ref28} Guan B, Zhao J. Affine Correspondences between Multi-Camera Systems for Relative Pose Estimation. IEEE Trans. Pattern Anal. Mach. Intell. 1-18, 2025.
\bibitem{ref29} Wu J, Guo X, Zhang D, et al. Numerical method for solving dynamical equations of multibody systems with nonholonomic constraints based on state-space method. Acta Mech. Sin. 42, 524764, 2026.
\bibitem{ref30} Guan B, Zhao J, Mitra S, et al. Six-Point Method for Multi-Camera Systems with Reduced Solution Space. Int. J. Comput. Vis. 133, 7270-7292, 2025.
\bibitem{ref31} Huang H, Chen C, Guan B, Tan Z, Shang Y, Li Z, Yu Q. Ridge estimation-based vision and laser ranging fusion localization method for UAVs. Appl. Opt. 64, 1352-1361, 2025.
\bibitem{ref32} Guan B, Zhao J, Kneip L. A Complete Solution to Generalized Relative Pose Estimation from Affine Correspondences. IEEE Trans. Pattern Anal. Mach. Intell. 1-15, 2026.




\end{thebibliography}
\end{document}